\documentclass[10pt,twocolumn,letterpaper]{article}

\usepackage{cvpr}              

\usepackage{subcaption}
\usepackage{graphicx}
\usepackage{amsmath}
\usepackage{amssymb}
\usepackage{booktabs}
\usepackage[super]{nth}

\usepackage{siunitx}
\newcommand{\pcent}[1]{\SI{#1}{\percent}}

 \newcommand{\thead}[1]{\small{\textsc{#1}}}

\usepackage[pagebackref,breaklinks,colorlinks]{hyperref}

\usepackage[capitalize]{cleveref}
\crefname{section}{Sec.}{Secs.}
\Crefname{section}{Section}{Sections}
\Crefname{table}{Table}{Tables}
\crefname{table}{Tab.}{Tabs.}

\def\cvprPaperID{17} 
\def\confName{CVPR}
\def\confYear{2022}

\begin{document}

\title{Automated Visual Monitoring of Nocturnal Insects \\ with Light-based Camera Traps}


\author{
	Dimitri Korsch $^1$ \hspace{.5cm}
	Paul Bodesheim $^1$ \hspace{.5cm}
	Gunnar Brehm $^2$ \hspace{.5cm}
	Joachim Denzler $^1$ \vspace{.2cm}\\
	$^1$ Computer Vision Group, \hspace{.5cm} $^2$ Phyletic Museum \\
	Friedrich Schiller University Jena, 07737 Jena, Germany \\
	{\tt\small \{dimitri.korsch, paul.bodesheim, gunnar.brehm, joachim.denzler\}@uni-jena.de}
}

\maketitle


\begin{abstract}
Automatic camera-assisted monitoring of insects for abundance estimations is crucial to understand and counteract ongoing insect decline.
In this paper, we present two datasets of nocturnal insects, especially moths as a subset of Lepidoptera, photographed in Central Europe.
One of the datasets, the EU-Moths dataset, was captured manually by citizen scientists and contains species annotations for \num{200} different species and bounding box annotations for those.
We used this dataset to develop and evaluate a two-stage pipeline for insect detection and moth species classification in previous work.
We further introduce a prototype for an automated visual monitoring system.
This prototype produced the second dataset consisting of more than \num{27000} images captured on \num{95} nights.
For evaluation and bootstrapping purposes, we annotated a subset of the images with bounding boxes enframing nocturnal insects.
Finally, we present first detection and classification baselines for these datasets and encourage other scientists to use this publicly available data.
\end{abstract}


\section{Introduction}
\label{sec:intro}

Climate change and the decline of species richness are severe challenges that influence the living conditions of humans around the world.
Especially the dramatic loss of insects~\cite{hallmann2017more,wagner2021insect} plays a crucial role in many ecological processes that affect agriculture and others.
Hence, monitoring insect species populations becomes more important nowadays to better understand insect decline and long-term trends in species distributions.
Furthermore, there are about one million named species on our planet~\cite{stork2018many}, making manual counting of individuals unrealistic.
Consequently, automated monitoring of insects is inevitably required to infer abundance estimations across larger regions.
One possible way is to use camera traps to collect images of insects that computer vision algorithms can then process to recognize the depicted species automatically.

In this paper, we focus on nocturnal insects, mainly nocturnal moths (Lepidoptera).
Even for this subset, there exist hundred thousands of different species worldwide and depending on the habitat, species lists can be narrowed down based on the study region.
For example, image datasets containing hundreds of moth species from Ecuador and Costa Rica are publicly available and can directly be used for evaluating fine-grained recognition algorithms~\cite{Rodner15:FRD}.
Here, we are interested in monitoring moth species in Central Europe.
We present datasets of moth images we have collected so far and our analysis of algorithms for insect localization and species classification.

\begin{figure}[t]
	\centering
	\includegraphics[width=0.95\linewidth]{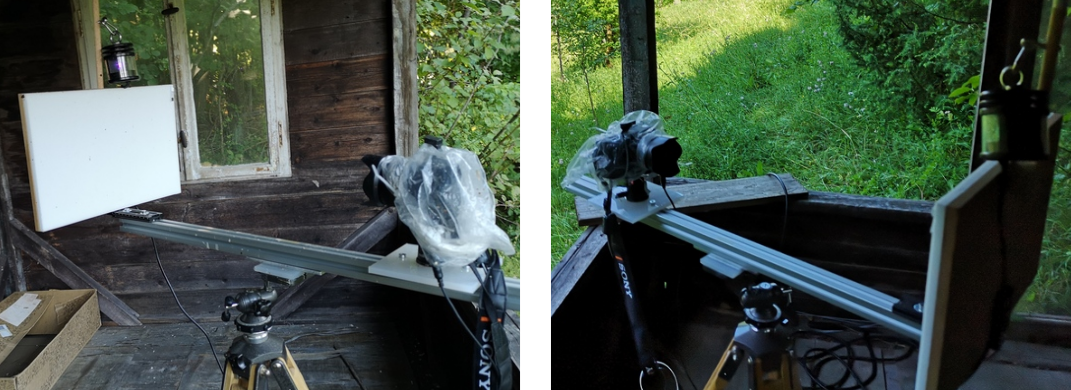}
	\caption{The prototype of the \emph{moth~scanner}: a white planar surface and an automated camera system.
	At night, UV light illuminates the surface to attract moths that land on the surface and the camera takes an image with flash every two minutes.}
	\label{fig:prototype}
\end{figure}

Our work is part of a larger project called AMMOD\footnote{\scriptsize{AMMOD = \textbf{A}utomated \textbf{M}ultisensor Station for \textbf{M}onitoring \textbf{o}f Bio\textbf{d}iversity (\url{https://ammod.de/})}}, which aims at developing self-sustaining multi-sensor stations for monitoring species diversity~\cite{Waegele22:TAM}.
One component of these stations is a light-based camera trap for nocturnal insects, called the \emph{moth~scanner}~\cite{Radig2021:AVL,Korsch21_DLP}.
It is a non-invasive monitoring system for automatically gathering images at nighttime.
A UV-LED lamp illuminates a white planar surface to attract the insects that land on this surface.
A high-resolution camera takes an image of the whole surface every two minutes.
Our prototype is shown in Figure~\ref{fig:prototype}.

With this setup, we can collect large-scale datasets of nocturnal insects over a long period that can then be used to develop and evaluate appropriate fine-grained species recognition algorithms.
The moth scanner takes several hundred images during one night, and within five months, we collected more than \num{27000} images with our prototype.
In this paper, we refer to the resulting dataset as the \emph{nocturnal insects dataset~(NID)}, and more details are given in Section~\ref{sec:dataset}.
Note that this dataset is supposed to be extended over time as our system will be in operation within the following years.
We plan to maintain multiple sensor stations in parallel at different locations.
Hence, it has the potential to become a valuable source for large-scale learning and continuous learning within a fine-grained domain.

Besides its impact on research in fine-grained recognition, our developments for automated visual monitoring of nocturnal insects are beneficial for ecologists.
Until now, insect monitoring is mainly done by hand and supported by citizen scientists who manually take images of individual insects in their gardens. 
Previously, we published an image dataset of nocturnal moths captured manually by citizen scientists, called \emph{\mbox{EU-Moths}} dataset at a local workshop~\cite{Korsch21_DLP}.
This paper also includes a dataset description and our baseline results for insect localization and species classification.
There are two reasons for this.
First, we want to announce this dataset to a broader audience interested in fine-grained recognition because it can directly be used for algorithm development and evaluation.
Second, we want to highlight the challenges for recognition algorithms that arise when processing automatically captured camera trap images compared to manually taken images with hand-held cameras.

In general, our paper aims to promote the application of moth species identification as a fine-grained visual recognition problem.
We underpin this with existing datasets, results of baseline algorithms, and a light-based camera trap setup that will be used during the following years to automatically collect further large-scale image data.
We believe that research on automated visual identification of hundreds to thousands of different nocturnal moth species can have a major impact on developing fine-grained recognition algorithms in general, and we, therefore, want to share our insights and datasets with the community.

\section{Related work}
\label{sec:related_work}

Besides insect monitoring~\cite{jonason2014lighttrap,svenningsen2020contrasting,bjerge2021automated,Radig2021:AVL,Korsch21_DLP}, there also exist automatic recognition systems for other animals.
They are often used to re-identify individuals of a certain species, e.g., great apes~\cite{Freytag16_CFW,Brust2017AVM,Kaeding18_ALR,yang2019great,sakib2020visual}, elephants~\cite{Koerschens18:Elephants,Koerschens19:ELPephants}, or sharks~\cite{hughes2017automated}, to name a few.
Of course, the field of fine-grained recognition mainly benefited from bird species datasets like CUB~\cite{WahCUB_200_2011} and NA-Birds~\cite{NABirds}.

For moth species identification, datasets exist with insects from Ecuador and Costa Rica containing images of 675 and 331 different species, respectively~\cite{Rodner15:FRD}.
There is only a single individual in each image spanning the whole image area.
In contrast to these datasets, we consider images of moth species from Central Europe recorded by light-based camera traps.
First, the recorded insects are still alive and may take various positions, making it harder to analyze the image.
Second, the captured image contains multiple insects, and individuals need to be localized before inferring the species.
Furthermore, we extend our datasets over time due to continuous monitoring.

There are also similar camera trapping systems developed by other groups, e.g., as reported in~\cite{bjerge2021automated}.
So far, they only consider eight different moth species and do not provide ground truth bounding box annotations for the individual insects.
In contrast, our EU-Moths dataset contains \num{200} different species, and we provide algorithms that can reliably recognize individuals from this larger set of species.


\section{The datasets}
\label{sec:dataset}

In the following, we present two datasets that we are using within our project for moth detection and species recognition.
They contain images of nocturnal insects, primarily moths, photographed on a white background.
In contrast to other publicly available datasets constructed for species classification such as iNaturalist~\cite{iNaturalist}, standardized camera trap images lead to a more homogeneous setting: the insects are photographed on a uni-color planar surface.

\begin{figure}[t]
	\centering
	\includegraphics[width=\linewidth]{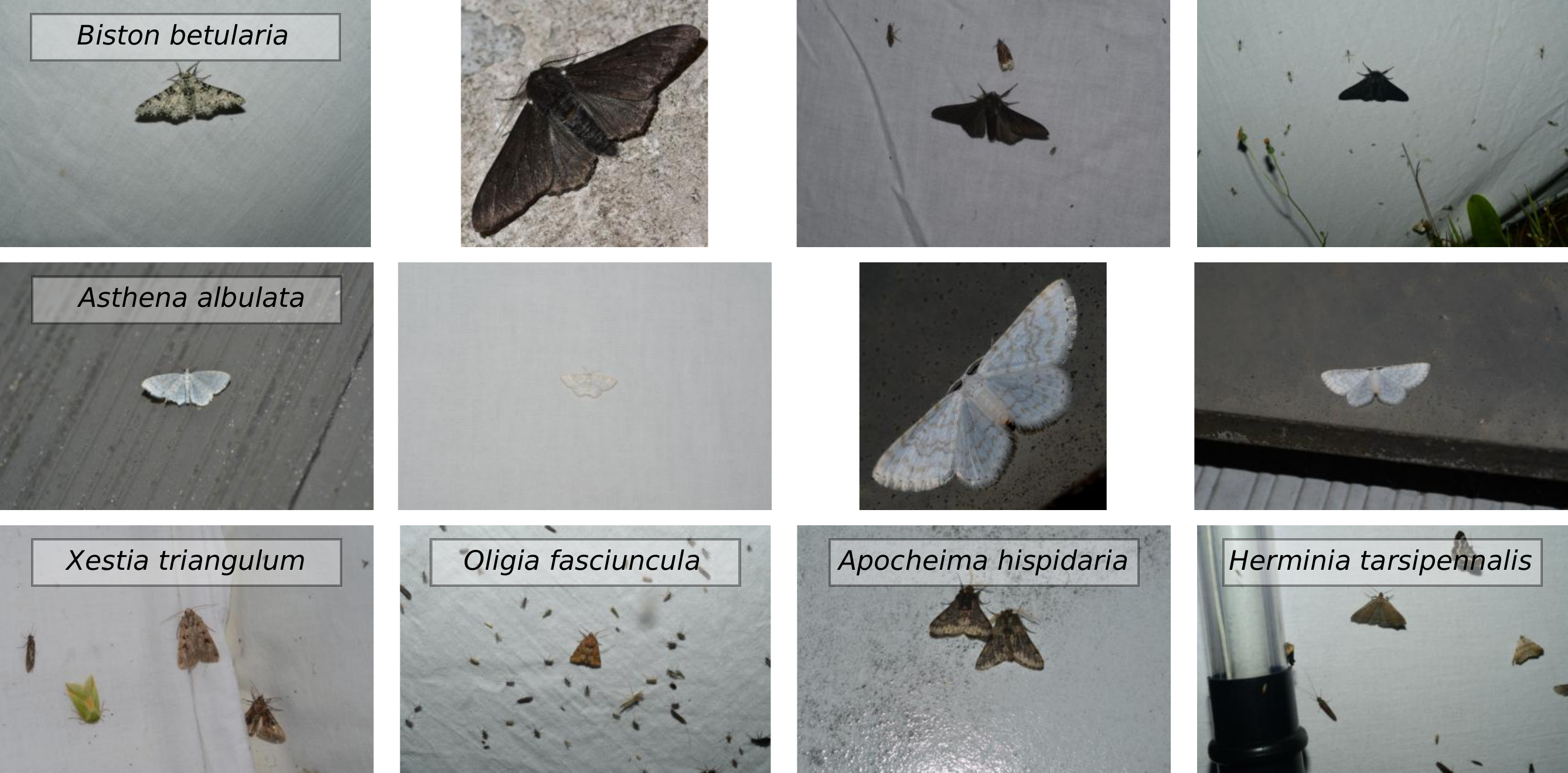}
	\caption{
		Example images from the \emph{EU-Moths} dataset.
		The first two rows show two different moths species, whereas the third row shows images with more than one insect.
		These examples illustrate the versatility in the appearance of the moths in the dataset.
	}
	\label{fig:examples:eumoths}
\end{figure}

\noindent\textbf{European moths (EU-Moths) dataset\footnote{\scriptsize{\url{https://inf-cv.uni-jena.de/eu_moths_dataset}}}:}\quad
This dataset consists of \num{200} species most common in Central Europe.
Each of the species is represented by approximately \num{11} images.
We consider a random but balanced split in eight training and three test images per species, resulting in roughly \num{1600} training and \num{600} test images in total.
Furthermore, we manually annotated bounding boxes for each insect.
For this dataset, citizen scientists photographed the insects manually and mainly on a relatively homogeneous background.
About \pcent{92} of the images contain only a single individual like it is shown in the first two rows of Figure~\ref{fig:examples:eumoths}.
The last row depicts images with more than one insect of interest.

\noindent\textbf{Nocturnal insects dataset (NID)\footnote{\scriptsize{\url{https://inf-cv.uni-jena.de/nid_dataset}}}:}\quad
Our camera trap setup takes high-resolution images of the insects resting on an illuminated surface.
We use a UV-LED lamp since this is the most attractive radiation for nocturnal insects than white light~\cite{brehm2017new}.
A 20-megapixel camera captured the images at an interval of two minutes (setup shown in Figure~\ref{fig:prototype}).

In five months (June - October \num{2021}), the system captured images during \num{95} nights, and we removed empty images without any insects at the beginning and the end of every night.
In total, we gathered more than \num{27000} images.

\begin{figure}[t]
	\centering
	\includegraphics[width=\linewidth]{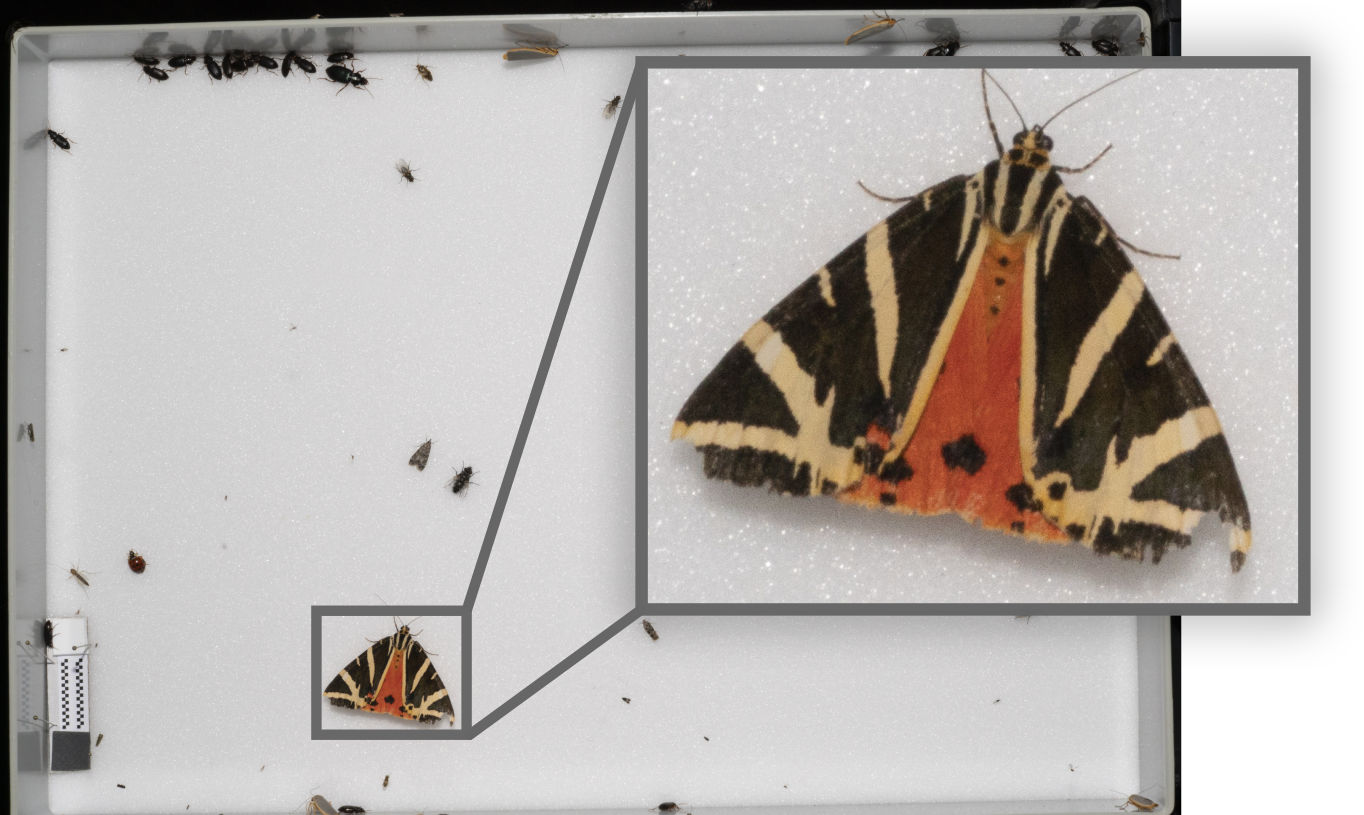}
	\caption{
		An example image of the \emph{NID} dataset.
		Two of the bounding boxes and the corresponding image patches are shown. 
		Note that even though these are relatively small parts of the original image, the visual species-related features in the patches have a high level of detail.
	}
	\label{fig:example}
\end{figure}

We first selected images from ten nights equally distributed over the entire period and manually annotated bounding boxes around insects in \num{818} images to evaluate detection methods.
As a result, we ended up with \num{9095} bounding box annotations.
Figure~\ref{fig:example} shows one of the images with two exemplary bounding boxes and the corresponding image patches.
In our baseline experiments, we use the first five nights for training and parameter tuning and data from the last five nights for the evaluation.

\section{Baseline methods}
\label{sec:methods}
As presented in preliminary work~\cite{Korsch21_DLP}, we deploy a two-stage pipeline for moth species detection and classification: (1) insect localization and (2) fine-grained moth species identification.
This separation is vital since the later prototypes will operate autonomously in the field and transmit the gathered images to central storage.
To reduce the amount of transmitted data, we will perform the detection directly at the moth scanner and transfer only the small image patches to the central storage.

\subsection{Single-shot detector}
\label{sub:meth:single_shot_detector}
We used a CNN-based state-of-the-art object detection model, namely the single-shot MultiBox detector (SSD) proposed by Liu~\etal~\cite{liu2016ssd}.
The authors utilize feature maps from multiple intermediate stages of the backbone CNN to predict location offsets and class confidences for a set of prior locations.
For more details about the loss functions, we refer to the original paper of Liu~\etal~\cite{liu2016ssd}.

\subsection{Fine-grained species classification}
\label{sec:meth:fgvc}

Neural networks, especially CNNs, yield state-of-the-art results in image classification tasks.
As we presented in our previous work~\cite{Korsch19_CSPARTS}, one can utilize a linear classifier with a sparsity-inducing L1-regularization to identify the most informative feature subsets of a high-dimensional (e.g., \num{2048} in case of InceptionV3) feature vector.
In combination with gradient maps~\cite{simonyan2013deep}, we use this subset of features to identify the regions of interest, the so-called saliency map, for an input image.
Afterward, we estimate with $k$-means clustering the spatial extent of coherent regions based on the identified saliency map and place bounding boxes around each region. 
The image patches of these bounding boxes serve as an unsupervised part representation, i.e., each region corresponds to a single part. 
These detected parts are finally used as additional input for the CNN classifier.
We refer to our previous work~\cite{Korsch19_CSPARTS,Korsch21_DLP} for more detail about the method and implementation details.


\section{Baseline results}
\label{sec:results}

We performed detection and classification experiments to produce the first baseline results on the presented datasets.
Species classification is only done on the EU-Moths dataset since the NID dataset has only bounding box annotations so far.
For evaluating the species classifier, we utilized the ground-truth bounding box annotations and only used the cropped image patches as inputs.

We repeated each experiment ten times and provided in Tables~\ref{tab:detection_results} and~\ref{tab:classification_results} the mean and standard deviation of the evaluation metrics across different runs.
We fine-tuned all models for \num{60} epochs and L2-regularization with a weight decay of~\num{5e-4}.
For both models, we utilized standard image augmentation methods: random cropping, random horizontal and vertical flipping, and color jittering (contrast, brightness, and saturation).

The SSD model was trained with an AdamW~\cite{AdamW} optimizer and a learning rate of \num{1e-3} for all epochs.
We used the VGG16~\cite{simonyan2014very} backbone architecture pre-trained on the ImageNet~\cite{ImageNet} with an input size of $300$px for the EU-Moths dataset and $512$px for the NID dataset.

The classification model was trained with an \mbox{RMSProp}~\cite{tieleman2012lecture} optimizer with an initial learning rate of \num{1e-4}, reduced by \num{0.1} after \num{20} and \num{40} epochs.
Further, we utilized label smoothing~\cite{inception} with a smoothing factor of \num{0.1}.
We used the InceptionV3 CNN architecture~\cite{inception} with the default input size of $299$px.
Additionally, we used two different pre-training methods.
Besides the typical ImageNet~\cite{ImageNet} pre-training, we used a pre-training on the iNaturalist~\cite{iNaturalist} dataset provided by Cui~\etal~\cite{Cui_2018_CVPR_large}.
Finally, we extract additional parts, as described in Sect.~\ref{sec:meth:fgvc}, and combine the predictions on these parts with the predictions on the entire image.

\subsection{Insect detection}
\label{sub:detection_results}

First, we report the detection performance on both datasets in Table~\ref{tab:detection_results} and we use mean average precision (mAP) as the evaluation metric.
The precision is computed based on two different intersection over union (IoU) thresholds of the predicted and the ground-truth bounding boxes.
The IoU-thresholds \num{0.5} and \num{0.75} (corresponding mAP denoted as \emph{mAP@0.50} and \emph{mAP@0.75}) are two typical choices used in the MC-COCO object detection benchmark~\cite{lin2014mscoco}.

\begin{table}[t]
	\begin{center}
	\begin{tabular}{lcc}
		\toprule
		 & \thead{mAP@0.75} & \thead{mAP@0.50} \\
		\midrule
		\thead{EU-Moths}
			& \num{88.88} \scriptsize{$(\pm \num{0.77})$}
			& \num{99.01} \scriptsize{$(\pm \num{0.09})$} \\
		\thead{NID Dataset}
			& \num{26.19} \scriptsize{$(\pm \num{5.64})$}
			& \num{91.21} \scriptsize{$(\pm \num{0.34})$} \\
		\bottomrule
	\end{tabular}
	\caption{
		Detection results on the EU-Moths and NID datasets.
	}
	\label{tab:detection_results}
	\end{center}
\end{table}

Based on the results, the SSD method performs significantly better on the EU-Moths dataset.
The baseline detector for the NID dataset achieves a stable mean-average precision of over \pcent{90} for the \emph{mAP@0.50} metric.
Nevertheless, the detector performs much worse for the more precise metric, the \emph{mAP@0.75}.
A possible explanation for this may be many small insects (as seen in Figure~\ref{fig:example}), where minor discrepancies between the prediction and ground truth degrade the results.
Even though we increased the input size for this dataset, the small sizes of some insects represent a challenge for the applied detection model.



\subsection{Species classification}
\label{sub:classification_results}

Table~\ref{tab:classification_results} shows the classification accuracies on the EU-Moth dataset for different setups.
First, we can observe the effect of the pre-training on different datasets.
Data used in the pre-training proposed by Cui~\etal~\cite{Cui_2018_CVPR_large} is more related to the domain of insects, and we can see this benefit in our reported results.

\begin{table}[t]
	\centering
	\begin{tabular}{lcc}
		\toprule
		 & \thead{ImageNet} & \thead{iNaturalist} \\		 
		 & \thead{pre-training} & \thead{pre-training} \\
		\midrule
		\thead{No Parts}

			& \num{89.46} \scriptsize{$(\pm \num{0.88})$}
			& \num{90.54} \scriptsize{$(\pm \num{1.10})$} \\
		\thead{With Parts}
			& \num{91.50} \scriptsize{$(\pm \num{0.61})$}
			& \num{93.13} \scriptsize{$(\pm \num{0.76})$} \\
		\bottomrule
	\end{tabular}
	\caption{
		Baseline classification results (accuracy in \%) on cropped images of the EU-Moths dataset.
	}
	\label{tab:classification_results}
\end{table}

Finally, utilizing methods for additional information extraction in the form of parts also improves the classification performance by approximately \pcent{2}, as Table~\ref{tab:classification_results} shows.
We achieved the best results using the part-based approach: \pcent{91.50} and \pcent{93.13} with ImagenNet and iNaturalist pre-training, respectively.


\section{Challenges of automated insect monitoring}
\label{sec:challenges}

As we worked with the images captured by our \emph{moth~scanner}, we faced some challenges that may interest others.
First, we obtained many images in a short period, and each image contains many insects of different sizes.
Annotating this huge amount of gathered data is very time-consuming.
Especially the manual species identification that requires expert knowledge is still ongoing, even though we use an annotation tool that supports with automatically inferred suggestions.
We further encountered typical challenges in real-world datasets like a long-tailed species distribution.

In our experiments, we also observed that current state-of-the-art detection models have problems with detecting tiny objects, as mentioned in Sect.~\ref{sub:detection_results}.
Although the uniform background might suggest that insect detection is an easy task in this scenario, fallen leaves and dirt on the surface and the large variability of insect sizes pose further challenges.
In contrast to the images recorded by citizen scientists with hand-held cameras that can focus on individual resting insects, automatically captured images of the light-based camera trap also contain motion blur and insects flying around that also (partially) occlude others.

Finally, the most challenging task is to perform the detection directly at the \emph{moth~scanner} (edge computing).
As mentioned previously in Sect.~\ref{sec:methods}, we plan to operate the camera traps autonomously in the field.
To reduce the amount of transmitted data, it is desirable to transfer only the small image patches instead of the entire image.
Unfortunately, this implies that current CNN-based state-of-the-art detection methods cannot be deployed on the hardware of the \emph{moth~scanner} with limited computational power and restricted energy budgets.
Hence, we will need to consider detection methods based on basic computer vision algorithms, like the blob detector presented by Bjerge~\etal~\cite{bjerge2021automated}.

\section{Conclusions}
\label{sec:conclusions}
In this paper, we presented a prototype of an automatic light-based camera trap for monitoring nocturnal insects.
The so-called \emph{moth~scanner} allows for capturing large-scale image datasets that can be used for moth localization and fine-grained species recognition.
Hence, this application domain can become an important area for research on fine-grained recognition with a large impact on ecology.
Besides the presented datasets, we also provided baseline results of a two-stage pipeline for detecting and classifying insects in images.






{\small
\bibliographystyle{ieee_fullname}
\bibliography{main}

\begin{thebibliography}{10}\itemsep=-1pt

\bibitem{bjerge2021automated}
Kim Bjerge, Jakob~Bonde Nielsen, Martin~Videbaek Sepstrup, Flemming
  Helsing-Nielsen, and Toke~Thomas H{\o}ye.
\newblock An automated light trap to monitor moths (lepidoptera) using computer
  vision-based tracking and deep learning.
\newblock {\em Sensors}, 21(2):343, 2021.

\bibitem{brehm2017new}
Gunnar Brehm.
\newblock A new led lamp for the collection of nocturnal lepidoptera and a
  spectral comparison of light-trapping lamps.
\newblock {\em Nota lepidopterologica}, 40:87, 2017.

\bibitem{Brust2017AVM}
Clemens-Alexander Brust, Tilo Burghardt, Milou Groenenberg, Christoph
  K{\"a}ding, Hjalmar K{\"u}hl, Marie Manguette, and Joachim Denzler.
\newblock Towards automated visual monitoring of individual gorillas in the
  wild.
\newblock In {\em ICCV Workshop on Visual Wildlife Monitoring (ICCV-WS)}, pages
  2820--2830, 2017.

\bibitem{Cui_2018_CVPR_large}
Yin Cui, Yang Song, Chen Sun, Andrew Howard, and Serge Belongie.
\newblock Large scale fine-grained categorization and domain-specific transfer
  learning.
\newblock In {\em Proceedings of CVPR}, 6 2018.

\bibitem{Freytag16_CFW}
Alexander Freytag, Erik Rodner, Marcel Simon, Alexander Loos, Hjalmar K{\"u}hl,
  and Joachim Denzler.
\newblock Chimpanzee faces in the wild: Log-euclidean cnns for predicting
  identities and attributes of primates.
\newblock In {\em German Conference on Pattern Recognition (GCPR)}, pages
  51--63, 2016.

\bibitem{hallmann2017more}
Caspar~A Hallmann, Martin Sorg, Eelke Jongejans, Henk Siepel, Nick Hofland,
  Heinz Schwan, Werner Stenmans, Andreas M{\"u}ller, Hubert Sumser, Thomas
  H{\"o}rren, et~al.
\newblock More than 75 percent decline over 27 years in total flying insect
  biomass in protected areas.
\newblock {\em PloS one}, 12(10):e0185809, 2017.

\bibitem{hughes2017automated}
Benjamin Hughes and Tilo Burghardt.
\newblock Automated visual fin identification of individual great white sharks.
\newblock {\em International Journal of Computer Vision}, 122(3):542--557,
  2017.

\bibitem{jonason2014lighttrap}
Dennis Jonason, Markus Franzén, and Thomas Ranius.
\newblock Surveying moths using light traps: Effects of weather and time of
  year.
\newblock {\em PLOS ONE}, 9(3):1--7, 03 2014.

\bibitem{Kaeding18_ALR}
Christoph K{\"a}ding, Erik Rodner, Alexander Freytag, Oliver Mothes, Bj{\"o}rn
  Barz, and Joachim Denzler.
\newblock Active learning for regression tasks with expected model output
  changes.
\newblock In {\em British Machine Vision Conference (BMVC)}, 2018.

\bibitem{Korsch19_CSPARTS}
Dimitri Korsch, Paul Bodesheim, and Joachim Denzler.
\newblock Classification-specific parts for improving fine-grained visual
  categorization.
\newblock In {\em Proceedings of the German Conference on Pattern Recognition},
  pages 62--75, 2019.

\bibitem{Korsch21_DLP}
Dimitri Korsch, Paul Bodesheim, and Joachim Denzler.
\newblock Deep learning pipeline for automated visual moth monitoring: Insect
  localization and species classification.
\newblock In {\em INFORMATIK 2021, Computer Science for Biodiversity Workshop
  (CS4Biodiversity)}, pages 443--460, 2021.

\bibitem{Koerschens18:Elephants}
Matthias K{\"o}rschens, Bj{\"o}rn Barz, and Joachim Denzler.
\newblock Towards automatic identification of elephants in the wild.
\newblock In {\em AI for Wildlife Conservation Workshop (AIWC)}, 2018.

\bibitem{Koerschens19:ELPephants}
Matthias K{\"o}rschens and Joachim Denzler.
\newblock Elpephants: A fine-grained dataset for elephant re-identification.
\newblock In {\em ICCV Workshop on Computer Vision for Wildlife Conservation
  (ICCV-WS)}, 2019.

\bibitem{lin2014mscoco}
Tsung-Yi Lin, Michael Maire, Serge Belongie, James Hays, Pietro Perona, Deva
  Ramanan, Piotr Doll{\'a}r, and C~Lawrence Zitnick.
\newblock Microsoft coco: Common objects in context.
\newblock In {\em ECCV}, pages 740--755. Springer, 2014.

\bibitem{liu2016ssd}
Wei Liu, Dragomir Anguelov, Dumitru Erhan, Christian Szegedy, Scott Reed,
  Cheng-Yang Fu, and Alexander~C Berg.
\newblock Ssd: Single shot multibox detector.
\newblock In {\em ECCV}, pages 21--37. Springer, 2016.

\bibitem{AdamW}
Ilya Loshchilov and Frank Hutter.
\newblock Decoupled weight decay regularization.
\newblock {\em arXiv preprint arXiv:1711.05101}, 2017.

\bibitem{Radig2021:AVL}
Bernd Radig, Paul Bodesheim, Dimitri Korsch, Joachim Denzler, Timm Haucke,
  Morris Klasen, and Volker Steinhage.
\newblock Automated visual large scale monitoring of faunal biodiversity.
\newblock {\em Pattern Recognition and Image Analysis}, pages 477--488, 2021.

\bibitem{Rodner15:FRD}
Erik Rodner, Marcel Simon, Gunnar Brehm, Stephanie Pietsch, J.~Wolfgang
  W{\"a}gele, and Joachim Denzler.
\newblock Fine-grained recognition datasets for biodiversity analysis.
\newblock In {\em CVPR Workshop on Fine-grained Visual Classification
  (CVPR-WS)}, 2015.

\bibitem{ImageNet}
Olga Russakovsky, Jia Deng, Hao Su, Jonathan Krause, Sanjeev Satheesh, Sean Ma,
  Zhiheng Huang, Andrej Karpathy, Aditya Khosla, Michael Bernstein, et~al.
\newblock Imagenet large scale visual recognition challenge.
\newblock {\em International journal of computer vision}, 115(3):211--252,
  2015.

\bibitem{sakib2020visual}
Faizaan Sakib and Tilo Burghardt.
\newblock Visual recognition of great ape behaviours in the wild.
\newblock {\em arXiv preprint arXiv:2011.10759}, 2020.

\bibitem{simonyan2013deep}
Karen Simonyan, Andrea Vedaldi, and Andrew Zisserman.
\newblock Deep inside convolutional networks: Visualising image classification
  models and saliency maps.
\newblock {\em arXiv preprint arXiv:1312.6034}, 2013.

\bibitem{simonyan2014very}
Karen Simonyan and Andrew Zisserman.
\newblock Very deep convolutional networks for large-scale image recognition.
\newblock {\em arXiv preprint arXiv:1409.1556}, 2014.

\bibitem{stork2018many}
Nigel~E Stork.
\newblock How many species of insects and other terrestrial arthropods are
  there on earth?
\newblock {\em Annual review of entomology}, 63:31--45, 2018.

\bibitem{svenningsen2020contrasting}
Cecilie~S Svenningsen, Diana~E Bowler, Susanne Hecker, Jesper Bladt, Volker
  Grescho, Nicole~M van Dam, Jens Dauber, David Eichenberg, Rasmus Ejrn{\ae}s,
  Camilla Fl{\o}jgaard, et~al.
\newblock Contrasting impacts of urban and farmland cover on flying insect
  biomass.
\newblock {\em bioRxiv}, 2020.

\bibitem{inception}
Christian Szegedy, Vincent Vanhoucke, Sergey Ioffe, Jon Shlens, and Zbigniew
  Wojna.
\newblock Rethinking the inception architecture for computer vision.
\newblock In {\em Proceedings of CVPR}, June 2016.

\bibitem{tieleman2012lecture}
Tijmen Tieleman and Geoffrey Hinton.
\newblock Lecture 6.5-rmsprop: Divide the gradient by a running average of its
  recent magnitude.
\newblock {\em COURSERA: Neural networks for machine learning}, 4(2):26--31,
  2012.

\bibitem{NABirds}
G. {Van Horn}, S. {Branson}, R. {Farrell}, S. {Haber}, J. {Barry}, P.
  {Ipeirotis}, P. {Perona}, and S. {Belongie}.
\newblock Building a bird recognition app and large scale dataset with citizen
  scientists: The fine print in fine-grained dataset collection.
\newblock In {\em Proceedings of the IEEE Conference on Computer Vision and
  Pattern Recognition}, pages 595--604, June 2015.

\bibitem{iNaturalist}
Grant Van~Horn, Oisin Mac~Aodha, Yang Song, Yin Cui, Chen Sun, Alex Shepard,
  Hartwig Adam, Pietro Perona, and Serge Belongie.
\newblock The inaturalist species classification and detection dataset.
\newblock In {\em Proceedings of the IEEE Conference on Computer Vision and
  Pattern Recognition}, pages 8769--8778, 2018.

\bibitem{wagner2021insect}
David~L Wagner, Eliza~M Grames, Matthew~L Forister, May~R Berenbaum, and David
  Stopak.
\newblock Insect decline in the anthropocene: Death by a thousand cuts.
\newblock {\em Proceedings of the National Academy of Sciences}, 118(2), 2021.

\bibitem{WahCUB_200_2011}
C. Wah, S. Branson, P. Welinder, P. Perona, and S. Belongie.
\newblock {The Caltech-UCSD Birds-200-2011 Dataset}.
\newblock Technical Report CNS-TR-2011-001, California Institute of Technology,
  2011.

\bibitem{Waegele22:TAM}
J.~Wolfgang Wägele, Paul Bodesheim, Sarah~J. Bourlat, Joachim Denzler, Michael
  Diepenbroek, Vera Fonseca, Karl-Heinz Frommolt, Matthias~F. Geiger, Birgit
  Gemeinholzer, Frank~Oliver Glöckner, Timm Haucke, Ameli Kirse, Alexander
  Kölpin, Ivaylo Kostadinov, Hjalmar~S. Kühl, Frank Kurth, Mario Lasseck,
  Sascha Liedke, Florian Losch, Sandra Müller, Natalia Petrovskaya, Krzysztof
  Piotrowski, Bernd Radig, Christoph Scherber, Lukas Schoppmann, Jan Schulz,
  Volker Steinhage, Georg~F. Tschan, Wolfgang Vautz, Domenico Velotto,
  Maximilian Weigend, and Stefan Wildermann.
\newblock Towards a multisensor station for automated biodiversity monitoring.
\newblock {\em Basic and Applied Ecology}, 2022.

\bibitem{yang2019great}
Xinyu Yang, Majid Mirmehdi, and Tilo Burghardt.
\newblock Great ape detection in challenging jungle camera trap footage via
  attention-based spatial and temporal feature blending.
\newblock In {\em Proceedings of the IEEE/CVF International Conference on
  Computer Vision Workshops}, 2019.

\end{thebibliography}
}

\end{document}